# A NOVEL METAHEURISTICS TO SOLVE MIXED SHOP SCHEDULING PROBLEMS

V. Ravibabu

Centre for Information Technology and Engineering,
Manonmaniam Sundaranar University, Tirunelveli, India

v_ravibabu@yahoo.com

*ABSTRACT*

*This paper represents the metaheuristics proposed for solving a class of Shop Scheduling problem. The Bacterial Foraging Optimization algorithm is featured with Ant Colony Optimization algorithm and proposed as a natural inspired computing approach to solve the Mixed Shop Scheduling problem. The Mixed Shop is the combination of Job Shop, Flow Shop and Open Shop scheduling problems. The sample instances for all mentioned Shop problems are used as test data and Mixed Shop survive its computational complexity to minimize the makespan. The computational results show that the proposed algorithm is gentler to solve and performs better than the existing algorithms.*

## 1. INTRODUCTION

Effective scheduling is an essential activity in manufacturing industry which leads to improvement in the efficiency and utilization of resources. This kind of problems in scheduling is called Shop Scheduling problems. A real world system may need the mixture of shop scheduling problems. In general, the Mixed Shop Scheduling problem is a NP-hard problem which is closely related and also the combination of the scheduling problems such as Job Shop, Flow Shop and Open Shop Scheduling problems. The Mixed Shop Scheduling problem can be assigned as n x m, where 'n' is the number of Jobs (J= {$j_1$, $j_2$... $j_n$}) can be processed on 'm' number of Machines (M= {$m_1$, $m_2$,...,$m_m$}). Each Job '$j_i$' consists of 'm' operations represented as '$M_{ij}$' processed on machine '$m_j$' for '$P_{ij}$' time units without pre-emption. One machine can process only one operation of same job at a time without interruption. The Mixed Shop Scheduling problem calculates the minimum makespan ($C_{max}$) for all operations in the order of n x m dimensions.

The Mixed Shop Scheduling problem is also called two production problems such as *Single-stage problem* and *Multi-stage problem*. A Single-stage system requires one operation for each job, whereas in a Multi-stage System there are jobs that require operations on different machines [2].

The Single-stage problem process on a single machine known as Single Machine Scheduling problem and the problem processing on more than one machines is known as the Parallel Machine Scheduling Problem. The Multi-stage problem is a Mixed Shop Scheduling problem which processes on three basic shop scheduling problems such as Job Shop, Flow Shop and Open Shop Scheduling Problems. In Job Shop Scheduling problem, the operation precedence constraint on the job is that the order of operations of job is fixed and the processing of an operation cannot be interrupted and concurrent. The machine processing constraint is that only a single job can be processed at the same time on the same machine. In Flow Shop Scheduling problem, each job is processed on machines in a fixed unidirectional order. In Open Shop Scheduling problem, each machine can process only one job at a time and each job can be processed by one machine at any time. Here, the routing of all operations is free.

## 2. RELATED WORKS

Yuri N. Sotskov, Natalia V.shakhlevich., surveyed the computational complexity of Mixed Shop Scheduling problem where the problem is a combination of Job Shop and Open Shop. Also the other classical scheduling assumptions were used for multi stage systems. This paper has no benchmark problems to discuss Mixed Shop problem in detail [1].

S.Q.Liu and H.L.Ong., presented the metaheuristics for the Mixed Shop Scheduling problem. In this paper, three metaheuristics were proposed for solving a class of basic shop and mixed shop scheduling problems. They proved that mixed shop is flexible compared to basic shop scheduling problems for 39 LA [01-39] benchmark instances [2].

Hanning Chen, Yunlong Zhu, and Kunyuan Hu., discussed a new variation, Cooperative Bacterial Foraging Optimization which improved the original Bacterial Foraging algorithm to solve the complex optimization problem. They solved for two cooperative approaches through the Bacterial Foraging Optimization algorithm, the serial heterogeneous on the implicit and hybrid space decomposition levels. This proposed method was compared with Particle Swarm Optimization, Bacterial Foraging Optimization and Genetic Algorithm [7].

Sambarta Dasgupta, Swagatam Das, Ajith Abraham, and Arijit Biswas (2009), represented a mathematical analysis of the chemotactics step in Bacterial Foraging Optimization from the viewpoint of the classical gradient descent search. They used two simple schemes for adapting the chemotactics step were height have been proposed and investigated. The adaptive variants of Bacterial Foraging Optimization were applied to the frequency-modulated sound wave synthesis problem [9].

E. Taillard [1989] has proposed a paper about Benchmarks' for Basic Scheduling Problems about 260 scheduling problems whose rare examples were published. Those kinds of problems correspond to real dimensions of industrial problems. In this paper he solved the flow shop, the job shop and the open shop scheduling problems and provides all benchmark results [10].

## 3. METAHEURISTICS

Metaheuristics makes assumptions on most of NP-Hard problems such as optimizing problems, decision making and search problems to be optimized and provides an unguaranteed optimal solution [15]. Optimization problems use computation method to search for an optimal solution and randomization method to get a least running time of the problem. The metaheuristics should be keen form of local search and starts with an initial solution. The search method should be efficient for Mixed Shop Scheduling Problem and satisfy the constraints by processing all operations without repetition.

## 3.1. Metaheuristics for Mixed Shop scheduling problem

The Bacterial Foraging Optimization algorithm is an evolutionary computation algorithm proposed by Passino.K.M (2002). This algorithm mimics the foraging behaviour of Escherichia Coli bacteria that living in human intestines. Escherichia Coli swims or moves by rotating its flagella on anti-clockwise direction and tumbles clockwise to choose new direction to swim for searching the nutrients. The foraging behaviour of Escherichia Coli process under three stages such as Chemotactics, Reproduction and Elimination and Dispersal Events. [12]

In Chemotactics, the health of the bacteria will be calculated by $j_{health}^{i} = \sum_{j=1}^{Nc+1} j(i,j,k,l)$

For $\theta^i(j,k,l)$, the bacteria 'i' undergoes $j^{th}$ chemotactics step by swimming and tumbling, the k is the reproduction step taken by total number of bacterium group 'S'. The group of bacteria arranged in terms of health and best half from the group is divided by two. The remaining bacteria will populated twice to make group in constant. The 'l' is the elimination and dispersal process which eliminates the rest of the population and dispersal makes random replacements in bacterium group.

The chemotactics process can be computed by $C_{(i)}$ while swimming and the run length of each bacteria can be calculated by $\theta^i(j+1,k,l) = \theta^i(j,k,l) + C_{(i)} \frac{\Delta(i)}{\sqrt{\Delta^T(i)\Delta(i)}}$, where $\Delta_{(i)}$ represents direction vector of the chemotactics.

The minimization of $J_{sw}^i = J^i + J_{cc}(\theta^i, \theta)$ can be calculated for bacteria's total cost value such that swarming function can be calculated by

$$J_{cc}(\theta^i, \theta) = \begin{cases} -M\left(\sum_{K=1}^{S} e^{-W_a \|\theta^i - \theta^k\|^2} - \sum_{K=1}^{S} e^{-W_r \|\theta^i - \theta^k\|^2}\right), with\ swarm \\ 0, without\ swarm \end{cases} \quad (1)$$

where M is the magnitude of the cell to cell signaling and $W_a$ and $W_r$ is the size of the attractant and repellent signals represented in Euclidean form [8].

The behavior of ant helps to explore the search space and each ant search for food at random by depositing pheromone in its path. The pheromone helps other ants to follow the same route. Each ant makes its own map. The pheromone value can be calculated through local update for single ant and global updates for group of ant at end of the tour.

The local pheromone represented by

$$\tau(r,s) \leftarrow (1-\rho).\tau(r,s) + \rho.\tau_0 \quad (2)$$

where $\tau$ denotes the pheromone lies between [0, 1].

Once all the ants reached their destination, the amount of pheromone values modified again and the global pheromone update represented by

$$\tau(r,s) \leftarrow (1-\rho).\tau(r,s) + \alpha.\Delta\tau(r,s) \quad (3)$$

33

Where $\Delta \tau(r,s) = \begin{cases} (L_{gb})^{-1}, & \text{if } (r,s) \in \text{global-best-tour} \\ 0, & \text{otherwise} \end{cases}$

Here denotes the pheromone decay parameter lies between [0, 1], $L_{gb}$ is the length of the globally best tour from the beginning of the trial and (r,s) is the pheromone addition on edge (r, s) [5].

Metaheuristics of ACO is to apply an ant tour repeatedly on all nodes to find the shortest path and this method also called stochastic greedy rule for optimal solution.[13]

$$s = \begin{cases} \arg\max_{u \in J(r)} \{[\tau(r,u)] \cdot [\eta(r,u)^\beta]\}, & \text{if } (q \leq q_0) \\ S, & \text{otherwise} \end{cases} \quad (4)$$

where (r,u) represents an edge between point r and u, and (r, u) stands for the pheromone on edge (r, u). (r, u) is the desirability of edge (r, u), which is defined as the inverse of the length of edge (r, u). q is a random number uniformly distributed in [0, 1], $q_0$ is a user-defined parameter lies between [0, 1] where q and $q_0$ is exploitation, is the parameter controlling the relative importance of the desirability. J (r) is the set of edges available at decision point r [5]. S is a random variable selected according to the probability distribution represented by

$$P(r,s) = \begin{cases} \dfrac{[\tau(r,u)] \cdot [\eta(r,u)^\beta]}{\sum_{u \in J(r)} [\tau(r,u)] \cdot [\eta(r,u)^\beta]}, & \text{if } (S \in J(r)) \\ 0, & \text{otherwise} \end{cases} \quad (5)$$

### 3.2. An Ant Inspired Bacterial Foraging Optimization Algorithm

**for** *Elimination-dispersal loop* **do**
  **for** *Reproduction loop* **do**
    **for** *Chemotaxis loop* **do**
      **for** *Bacterium i* **do**
        **Tumble**: Generate a secure random vector q $\in$ decimal.
        **If** q < $q_0$ **then**
          Generate a secure random vector l $\in$ operation,
          ph[job][operation] based on equation 4.
        **Else**
          Generate a secure random vector l $\in$ operation,
          ph[job][operation] based on equation 5.
          **end**
        **Move**: Generate a secure random vector $l_{new}$ $\in$ operation.
         **Swim**:
         **if** time[job][l] < time[job][$l_{new}$] **then**
          current_operation = l
         **Else**
          current_operation = $l_{new}$
        **end**
      **end**

```
      end
    end
    Sort bacteria in J_st.
    S_r = S/2 bacteria with the highest J value die,
    S_r bacteria with the updated value of J and J_st .
  end
  Eliminate and disperse with p_ed.
   Update J and J_st.
End
```
*Note: Refer Appendix for Nomenclature*

## 4. IMPLEMENTATION

The Mixed Shop Scheduling problem can be represented in terms of O = n x m where O is the number of nodes and n denotes the job and m denotes the machines. The $P_{ij}$ is processing time of operations.

Table 1. Mixed Shop Problem

|       | M1 | M2 | M3 |
|-------|----|----|----|
| $J_J$ | 2  | 2  | 3  |
| $J_F$ | 1  | 3  | 2  |
| $J_O$ | 3  | 1  | 2  |

The sample test problem generated by 3 jobs and 3 machines such that the completion time based on Mixed Shop scheduling problem, the mixture of Job Shop, Open Shop and Flow Shop were shown in table 1. Each job is allocated based on the constraints of each Shop Scheduling problems such as $J_J$, $J_F$ and $J_O$ is apportioned with Job Shop, Flow Shop and Open Shop. The computational result shows that the makespan is minimum for the Mixed Shop Scheduling Problem is 7 were shown in table 2, where remaining shop problems completes its process with makespan value of 11, 11 and 8.

Table 2. Result for Mixed Shop Scheduling Problem

| M3 | $O_3$ |        | $J_3$ |   |   | $F_3$ |   |
|----|-------|--------|-------|---|---|-------|---|
| M2 | $J_2$ |        | $F_2$ |   |   | $O_2$ | ------ |
| M1 | $F_1$ | ------ | $O_1$ |   |   | $J_1$ |   |
| 0  | 1     | 2      | 3     | 4 | 5 | 6     | 7 |

The Mixed Shop Scheduling problem has constraints of Job Shop, Flow Show and Open Shop for each machines and obtained minimum completion time is 7 where Open Shop Scheduling problem has no restrictions in its operations and obtained minimum completion time is 8. The Mixed Shop Scheduling problem achieves better minimum makespan in Hybrid Bacterial Foraging Optimization algorithm when compared to Job Shop, Flow Shop and Open Shop Scheduling problems. The same techniques were used to test the complexity of Mixed Shop by

35

International Journal in Foundations of Computer Science & Technology (IJFCST), Vol. 3, No.2, March 2013

computing it for all relevant benchmark problems. Since there were no direct instances for Mixed Shop Scheduling, the sample instances (RND) were generated under the constraints of Shop Scheduling problems and used as test data for Job Shop, Flow Shop, Open Shop and Mixed Shop Scheduling problems.

The RND instances were implemented based on the Bacterial Foraging Algorithm (BFO) were shown in table 3 and the figure 1 indicates the results of table 3 in chart representation. The Bacterial Foraging Optimization Algorithm was compared for Job Shop, Flow Shop, Open Shop and Mixed Shop Scheduling problem were the results are not much comparable. So the Bacterial Foraging Optimization Algorithm is altered according to the shortest path selection method and implemented as an Ant Inspired Bacterial Foraging Optimization Algorithm (ABFO) for Job Shop, Flow Shop, Open Shop and Mixed Shop Scheduling problems were shown in table 4 and the figure 2 indicates the results of table 4 in chart representation.

Table 3. Comparison Result for Job Shop, Flow Shop, Open Shop and Mixed Shop Scheduling Problems using BFO

| INSTANCE [SIZE] | JOB SHOP | FLOW SHOP | OPEN SHOP | MIXED SHOP |
|---|---|---|---|---|
| RND [3X3] | 310 | 310 | 240 | 240 |
| RND [5X5] | 495 | 515 | 350 | 345 |
| RND [7X7] | 720 | 727 | 690 | 672 |
| RND[10X10] | 1240 | 1320 | 890 | 810 |
| RND[15X15] | 1720 | 1810 | 1540 | 1415 |
| RND[20X20] | 2814 | 2730 | 2240 | 1940 |

Table 4. Comparison Result for Job Shop, Flow Shop, Open Shop and Mixed Shop Scheduling Problems ABFO

| INSTANCE [SIZE] | JOB SHOP | FLOW SHOP | OPEN SHOP | MIXED SHOP |
|---|---|---|---|---|
| RND [3X3] | 285 | 285 | 201 | 201 |
| RND [5X5] | 452 | 452 | 313 | 305 |
| RND [7X7] | 664 | 664 | 481 | 469 |
| RND[10X10] | 1045 | 1080 | 764 | 749 |
| RND[15X15] | 1653 | 1662 | 1213 | 1155 |
| RND[20X20] | 2314 | 2343 | 1721 | 1672 |

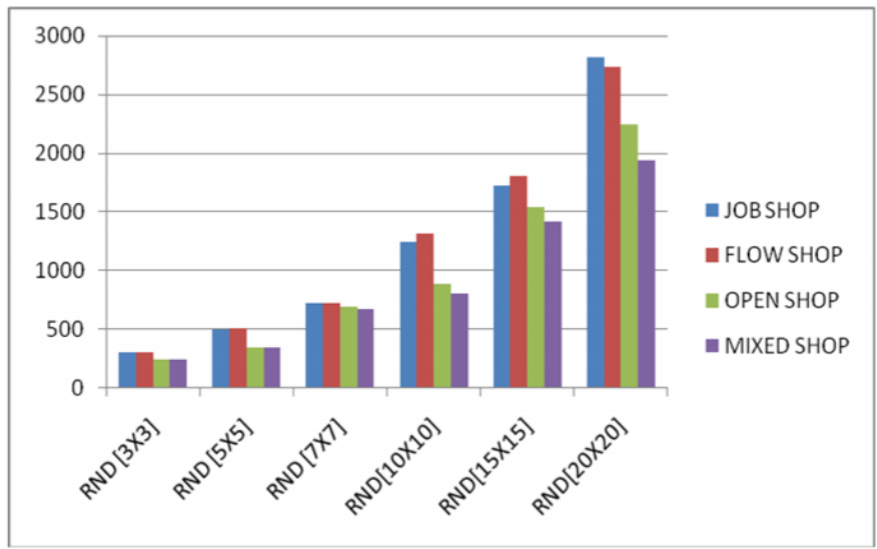

Figure 1. Comparison Result for Job Shop, Flow Shop, Open Shop and Mixed Shop Scheduling Problems BFO

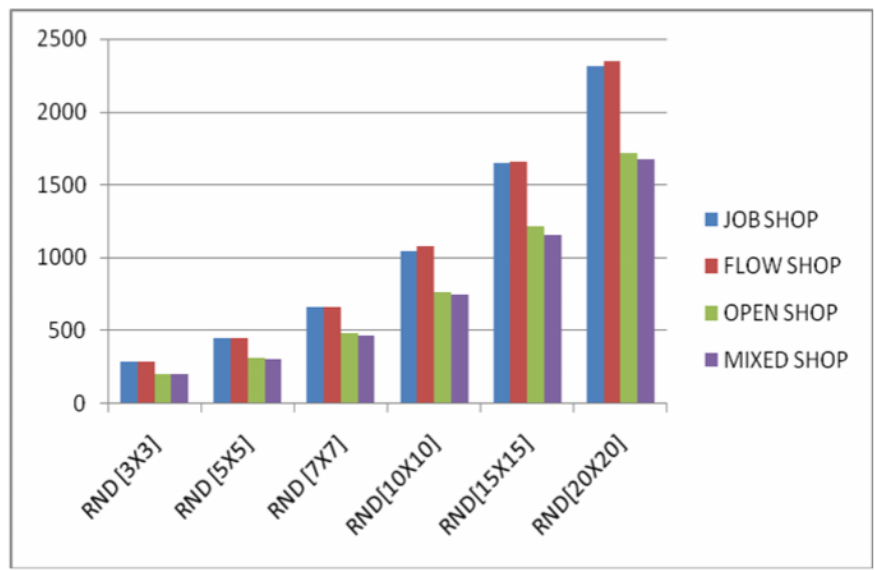

Figure 2. Comparison Result for Job Shop, Flow Shop, Open Shop and Mixed Shop Scheduling Problems using ABFO

The implementation was done in Java 6.0; the Pseudo Random Number Generator (PRNG) is used to provide secure random class for choosing next node in search space. The constant values were used as parameter to check the performance of the problems. The Mixed Shop Scheduling Problem achieves the best optimum values of the all other shop scheduling problems using Ant Inspired Bacterial Foraging Algorithm.

37

## 5. CONCLUSIONS

In computational experiments the performance of Mixed Shop Scheduling problem was evaluated by applying the sample test data of the basic shop scheduling problems and the metaheuristics shows that the proposed Ant Inspired Bacterial Foraging Optimization algorithm performs well on Mixed Shop Scheduling problem and can achieve minimum makes span than Job Shop and Flow Shop. The Open shop scheduling has been achieved to its best optimum value and also the same instance implemented on Mixed Shop problem achieved the least optimum values for most instances. However, the Ant Inspired Bacterial Foraging Optimization Algorithm performs well on all these kind of shop scheduling problems.

## 6. APPENDIX

**NOMENCLATURE**

| | | |
|---|---|---|
| $C_{(i)}$ | - | Step size |
| i | - | Bacterium number |
| j | - | Counter for chemotactic step |
| J(i, j, k, l) | - | Cost at the location of $i^{th}$ bacterium |
| $J_{cc}$ | - | Swarm attractant cost |
| $J^i_{health}$ | - | Health of bacteria |
| $J^i_{sw}$ | - | Swarming effect |
| k | - | Counter for reproduction step |
| l | - | Counter for elimination-dispersal step |
| $N_c$ | - | Maximum number chemotactic steps |
| $N_{ed}$ | - | Number of elimination dispersal event |
| $N_{re}$ | - | Maximum reproduction steps |
| Ns | - | Maximum number of swims |
| P | - | Dimension of the optimization |
| $P_{ed}$ | - | Probability of occurrence of Elimination-dispersal events |
| S | - | Population of the E. coli bacteria |
| $^i(j, k, l)$ | - | Location of the $i^{th}$ bacterium at $j^{th}$ chemotactic step, $k^{th}$ reproduction step, and $l^{th}$ elimination-dispersal step |
| $J_J$ | - | Job Shop Constraints |
| $J_F$ | - | Flow Shop Constraints |
| $J_O$ | - | Open Shop Constraints |

## REFERENCES

[1] Yuri N. Sotskov, Natalia V. Shakhlevich, "Mixed Shop Scheduling Problems,' INTAS (project 96 0820) and ISTC (project B 104 98).

[2] S. Q. Liu and H. L. Ong, "Metaheuristics for the Mixed Shop Scheduling Problem," Asia-Pacific Journal of Operational Research, Vol. 21, No. 4, 2004, pp. 97-115.

[3] W. J. Tang, Q. H. Wu, and J. R. Saunders, "Bacterial Foraging Algorithm For Dynamic Environments," IEEE Congress on Evolutionary Computation, July 2006, pp. 16-21.

[4] Hai Shen et al, "Bacterial Foraging Optimization Algorithm with Particle Swarm Optimization Strategy for Global Numerical Optimization,"GEC'09, June 2009.

[5] Jun Zhang, Xiaomin Hu, X.Tan, J.H Zhong and Q. Huang., "Implementation of an Ant Colony Optimization Technique for Job Shop Scheduling Problem," Transactions of the Institute of Measurement and Control 28, pp. 93_/108, 2006.
[6] Peter Bruker, "Scheduling Algorithms," Fifth Edition, Springer-Verlag Berlin Heidelberg, 2007.
[7] Hanning Chen, Yunlong Zhu, and Kunyuan Hu ., "Cooperative Bacterial Foraging Optimization," Hindawi Publishing Corporation, Discrete Dynamics in Nature and Society, Article ID 815247, Volume 2009.
[8] Jing Dang, Anthony Brabazon, Michael O'Neill, and David Edition., "Option Model Calibration using a Bacterial Foraging Optimization Algorithm", LNCS 4974, 2008.
[9] Sambarta Dasgupta, Swagatam Das, Ajith Abraham, Senior Member, IEEE, and Arijit Biswas., "Adaptive Computational Chemotaxis in Bacterial Foraging Optimization: An Analysis," IEEE Transactions on Evolutionary Computation, vol. 13, no. 4, August 2009.
[10] E. Taillard, "BenchMarks For Basic Scheduling Problems," European Journal of Operations Research, 64, 1993, pp. 278-285.
[11] Jason Brownlee., "Clever Algorithms: Nature-Inspired Programming Recipes," First Edition, January 2011.
[12] Kevin M. Passino., "Bacterial Foraging for Optimization," International Journal of Swarm Intelligence Research, 1(1), January – March, 2010, pp. 1-16.
[13] Ravibabu. V, Amudha. T, "An Ant Inspired Bacterial Foraging Methodology Proposed To Solve Open Shop Scheduling Problems", International Journal of Advanced Research in Computer Science and Electronics Engineering (IJARCSEE), ISSN: 2277-9043, August 2012.
[14] J. E. Beasley, OR-Library Web page for Instances, http://people.brunel.ac.uk/~mastjjb/jeb/info.html
[15] Naoyuki Tamura, CSP2SAT: Open Shop Scheduling Problems Web page, http://bach.istc.kobe-u.ac.jp/csp2sat/oss/
[16] http://www.metaheuristics.org/

**Authors**

**Mr.V.Ravibabu** received his B.Sc Degree in Computer Science and MCA Degree in Computer Applications in 2009 and 2012 respectively, from Bharathiar University, Coimbatore, India. His area of interest includes Agent based computing and Bio-inspired computing. He has attended National / International Conferences and 1 research publication for his credit in International Journal. He is a member of International Association of Engineers.

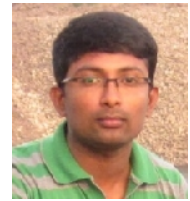